%% file: main.tex
\documentclass[letterpaper]{article} 
\usepackage{aaai23}  
\usepackage{times}  
\usepackage{helvet}  
\usepackage{courier}  
\usepackage[hyphens]{url}  
\usepackage{graphicx} 
\usepackage{natbib}  
\usepackage{caption} 
\usepackage{amssymb,amsthm}
\usepackage{amsmath}
\usepackage{amsfonts}
\usepackage{mathrsfs}
\usepackage{tabularx} 
\usepackage[linesnumbered,boxed,ruled,commentsnumbered]{algorithm2e}

\usepackage{subfigure}
\usepackage{acronym}
\usepackage{todonotes}
\usepackage{color}
\usepackage{mathtools}
\usepackage{booktabs}
\usepackage{multirow}
\input{defs.tex}

\newcommand{\defeq}{\mathrel{\mathop:}=}

\title{Shared Information-Based Safe And Efficient Behavior Planning \\ For Connected Autonomous Vehicles}
\author {
    Songyang Han\textsuperscript{\rm 1},
    Shanglin Zhou\textsuperscript{\rm 1},
    Lynn Pepin\textsuperscript{\rm 1},
    Jiangwei Wang\textsuperscript{\rm 2},
    Caiwen Ding\textsuperscript{\rm 1},
    Fei Miao\textsuperscript{\rm 1} 
}
\affiliations {
    \textsuperscript{\rm 1} Department of Computer Science and Engineering, University of Connecticut\\
    \textsuperscript{\rm 2} Department of Electrical and Computer Engineering, University of Connecticut\\
    \{songyang.han, shanglin.zhou, lynn.pepin, jiangwei.wang, caiwen.ding, fei.miao\}@uconn.edu
}

\begin{document}

\maketitle

\begin{abstract}
The recent advancements in wireless technology enable connected autonomous vehicles (CAVs) to gather data via vehicle-to-vehicle (V2V) communication, such as processed LIDAR and camera data from other vehicles. In this work, we design an integrated information sharing and safe multi-agent reinforcement learning (MARL) framework for CAVs, to take advantage of the extra information when making decisions to improve traffic efficiency and safety. We first use weight pruned convolutional neural networks (CNN) to process the raw image and point cloud LIDAR data locally at each autonomous vehicle, and share CNN-output data with neighboring CAVs. We then design a safe actor-critic algorithm that utilizes both a vehicle's local observation and the information received via V2V communication to explore an efficient behavior planning policy with safety guarantees. Using the CARLA simulator for experiments, we show that our approach improves the CAV system's efficiency in terms of average velocity and comfort under different CAV ratios and different traffic densities. We also show that our approach avoids the execution of unsafe actions and always maintains a safe distance from other vehicles. We construct an obstacle-at-corner scenario to show that the shared vision can help CAVs to observe obstacles earlier and take action to avoid traffic jams. 
\end{abstract}

\section{Introduction}
Wireless communication technologies such as WiFi and 5G cellular networks help enable vehicle-to-vehicle (V2V) communication. The U.S. Department of Transportation (DOT) estimated that DSRC (Dedicated Short-Range Communications)-based V2V communication could potentially address up to 82\% of crashes in the U.S. every year~\cite{orosz2017seeing, DSRC_standard}. Basic safety messages (BSMs) sharing benefits connected autonomous vehicles (CAVs) coordination for intersections and lane-merging~\cite{Coordinate_CAV, CV_intersection, ort2018autonomous}.

Vision information captured by the onboard cameras and LIDARs can also improve decision making. Shared vision information among vehicles provides a see-through view to human drivers for reactive early lane changing~\cite{kim2015impact}, reduces the uncertainty due to blind spots of cooperative trajectory planning~\cite{buckman2020generating}. 
However, when CAVs get extra environment knowledge via V2V communication, how to make prudent decisions to improve traffic efficiency, whether sharing vision information can bring benefits are still unsolved challenges. Hence, we show how CAVs can take advantage of information sharing to make better driving decisions while meeting safety guarantees and improving traffic efficiency.




In this work, we design a safe actor-critic algorithm integrated with information sharing to enhance operational performance. To save computing resources, we first develop a convolutional neural network (CNN) weight pruning technique to process the camera and LIDAR data, and share the output with neighboring CAVs. Then the raw images and point clouds are synchronously processed and shared as the MARL input. The mixed traffic driving environment includes autonomous and human-driving vehicles. We model the behavior planning problem as a decentralized partially observable Markov decision process (Dec-POMDP)~\cite{oliehoek2016concise}. We design a safe action mapping algorithm to make sure that the actions trained and executed by our proposed MARL algorithm are safe. 
In experiments, we show that our approach increases traffic efficiency and guarantees safety.

In summary, the main contributions of this work are:
\begin{itemize}
    \item We design an integrated information sharing and safe multi-agent reinforcement learning framework to utilize the shared information via V2V for the behavior planning of CAVs to improve traffic efficiency and safety. 
    \item We implement the safe actor-critic in a simulator CARLA that can simulate multiple vehicles with physical dynamics. The experiment shows the designed framework can improve average velocity and comfort for CAVs with safety guarantees, for CAV-only scenario and mixed traffic environment that includes both autonomous and human driven vehicles. Our result also gives insight that the traffic flow and comfort can be improved when the CAV's penetration arises. 
    \item We validate our integrated multi-agent reinforcement learning framework in challenging driving scenarios like obstacle-at-corner, and the shared vision can help vehicles avoid obstacles in advance.
\end{itemize}


\section{Related Work}
\paragraph{Deep Learning For Autonomous Vehicles}
It is productive for autonomous vehicles to learn the steering angle and acceleration control directly based on vision input, such as end-to-end imitation learning~\cite{le2022survey}, and end-to-end reinforcement learning~\cite{cheng2019end,chen2020conditional,chen2021interpretable}, but they cannot guarantee safety. The other popular way is to separate the learning and control phases~\cite{shalev2016safe,aradi2020survey,he2021rule,zhang2022unified}. . The learning methods can give a high-level decision making, such as ``go straight", ``go left"~\cite{pan2017virtual}, whether yield to another vehicle or not~\cite{shalev2016safe}. It can also extract image features and then apply control upon these features~\cite{chen2015deepdriving}. However, the works mentioned above do not consider the connection between CAVs, while we consider how CAVs use information sharing to improve safety and efficiency.

\paragraph{Multi-agent Reinforcement Learning}
Existing multi-agent reinforcement learning (MARL) literature~\cite{zhang2019multi} has not fully solved the challenges for CAVs. How communication among agents will improve systems' safety or efficiency in policy learning has not been addressed. Recent advancements like multi-agent deep deterministic policy gradient (MADDPG)~\cite{lowe2017multi}, the attention mechanism~\cite{iqbal2019actor}, cooperative MARL~\cite{rashid2018qmix,sunehag2018value,yu2019distributed,rashid_nips20,NEURIPS2020_8977ecbb} and league training~\cite{vinyals2019grandmaster} do not specify communication among agents or safety guarantee for the learned policy. We consider a novel problem with information sharing, and design a safe actor-critic algorithm with centralized training and decentralized execution~\cite{foerster2017counterfactual, rashid2018qmix}. 

\paragraph{Safe Reinforcement Learning}
Safe reinforcement learning is an increasingly important research area for real world safety-critical applications~\cite{berkenkamp2017safe, bastani2018verifiable, alshiekh2018safe, fisac2019bridging, cheng2019end,thomas2021safe}, including model predictive control (MPC)-based supervised learning~\cite{chen2018approximating} and guided policy search for the DRL~\cite{MPCGPS_icra16}, joint learning of control policy and control barrier certificates~\cite{qin2021learning}. Optimization-based methods are proposed to prove safety of the learned policy~\cite{berkenkamp2017safe, bastani2018verifiable, fazlyab2020safety, fazlyab2019efficient}. However, they are not directly applicable to multi-agent systems with complicated dynamic environment such as CAVs. Safe MARL methods mainly have two types: constrained MARL or shielding for exploration. In constrained MARL~\cite{lu2021decentralized}, agents maximize the total expected return while keeping the costs lower than designed bounds. However, the constraints in these methods cannot explicitly represent the safety requirement at every timestep of a physical dynamic system like CAVs, and safety is rarely guaranteed during learning in practice. The model predictive shielding (MPS) algorithm provably guarantees safety for any learned MARL policy~\cite{li2020robust, zhang2019mamps}. The basic idea is to use a backup controller to override the learned policy by dynamically checking whether the learned policy can maintain safety. 
However, this overriding interrupts the learning process. Our proposed algorithm maps any action in the action space to a safe action such that the RL agent can keep learning without being interrupted.

\section{Problem Description}
\label{sec:problem}
This work addresses how to utilize shared information to make better behavior decisions such as when to change/keep lane for CAVs, considering safety guarantees and the improvement of the transportation system efficiency. The environments can be mixed with CAVs and human-driven vehicles, and our decision making framework is designed for CAVs. Human-driven vehicles in the environment can be observed by the sensors on autonomous vehicles and road infrastructures. The V2V communication provides environment information to a single vehicle beyond what its own sensors can detect. We formulate this problem as a Dec-POMDP \cite{oliehoek2016concise}: 


\begin{definition}[\textbf{Decentralized Partially Observable Markov Decision Process (Dec-POMDP)}] A Dec-POMDP is a collection $G = (\mathcal{S}, \mathcal{N}, \mathcal{A}, \mathcal{O}, \mathcal{T}, \mathbf{o}, \mathbf{r}, n)$: set $\mathcal{S}$ is a set of states of the world; $\mathcal{N} = \{1, ..., n\}$ is the set of agents; $\mathcal{A}$ is the joint action set, and the joint action is $ a = (a^1, ..., a^n) \in \mathcal{A} \defeq \mathcal{A}^1 \times \cdots \times \mathcal{A}^n$; $\mathcal{O}$ is the observation set; $\mathcal{T}: \mathcal{S} \times \mathcal{A} \times \mathcal{S} \rightarrow [0,1]$ is the state transition function; all agents share the same reward function $\mathbf{r}: \mathcal{S} \times \mathcal{A} \rightarrow \mathbb{R}$; $\mathbf{o}: \mathcal{S} \times \mathcal{A} \rightarrow \mathcal{O}$ is the observation function that outputs the observation each individual agent receives.
\end{definition}
\begin{definition}[\textbf{Safe action}] 
An action is said to be safe for an ego vehicle $i$ if taking this action, the distance between $i$ and all its ``one hop" neighbors $j$ in the forward direction is above a safe distance $\mathcal{D}_s$, regardless of whether this neighbor is a CAV or not. That is to say, $\norm{p_i - p_j} > \mathcal{D}_s$ where $p_i$ and $p_j$ represent the spatial positions of vehicle $i$ and $j$. An action is a safe action if it is safe for all the time steps until the vehicle stops or changes to a different action.
\end{definition}
We want to find a policy $\pi(a|o)$ in the Dec-POMDP to maximize the total expected return $R_t = \sum_{k = 0}^{ \infty } \gamma^{k} r_{t+k}$ where $\gamma$ is a discount factor, and make sure there exists a controller to execute the action safely. Hence, we propose a safe actor-critic algorithm to learn the action value function $Q(o, a)$ with a safe action mapping $\Phi$ to guarantee policies are only explored in the safe action space. The details of this algorithm will be explained in Section~\ref{sec:learning}. The policy to generate the behavior planning decisions $\pi(a | o)$ should maximize the total expected return while maintaining $a$ safe.

We use the centralized training and decentralized execution paradigm~\cite{ foerster2017counterfactual, rashid2018qmix}, a common design of the MARL. 
The training process can make full use of all agents' information while each agent can only access its own observation in the execution. We assume all CAVs are cooperating with each other, and a global reward is usually assigned to each agent as the cooperative MARL examples shown in \cite{iqbal2019actor, rashid2018qmix}. To limit the scope of this paper, we use one global reward for all the agents. Credit assignment can be included to further improve the result like the methods in \cite{sunehag2018value, foerster2017counterfactual}.

We consider a partially observable environment with unknown state transition $\mathcal{T}$ and unknown observation function $\mathbf{o}$ in our work. This problem impedes using value function based RL methods. One practical approach to solve it is to replace the observation by the past action-observation history $h_t = o_1 a_1, \ldots, o_{t-1} a_{t-1}, o_{t}$ to construct a Markovian observation such that we can apply RL methods. 
This practical approach shows good performance in the literature~\cite{sunehag2018value, hausknecht2015deep}. Hence, in this paper, we include the past action-observation history in the $Q$-network. 

\section{Information Sharing}
\label{sec:info_sharing}



The ego vehicle's observation has three sources: (i) its own onboard sensors, e.g., inertial measurement unit (IMU), camera, and LIDAR; (ii) shared actions of neighboring vehicles; (iii) the shared vision information that includes features \{lane index, distance, observation angle, rotation\} of neighboring vehicles provided by the processed vision (shown in Fig.~\ref{vision_plot}) of its immediate leaders in its current lane and neighbor lanes, via V2V communication. 

\begin{definition}[\textbf{Immediate leader and follower}]
One vehicle $i$ is called the immediate follower of a vehicle $j$ on lane \#$k$ if $ i = \mathop{\arg\min}_{i} \  \norm{ x_i- x_j } \ s.t. \ x_i<x_j, l_i = k$, where $i$ and $j$ are vehicles' indexes, $x_i$ and $x_j$ represent the longitudinal positions of vehicle $i$ and $j$, $l_i$ is vehicle $i$'s lane number. Vehicle $j$ is called the immediate leader of $i$ on lane \#$k$.
\end{definition}
\begin{figure}[h]
  \centering
  \includegraphics[width=3.2in]{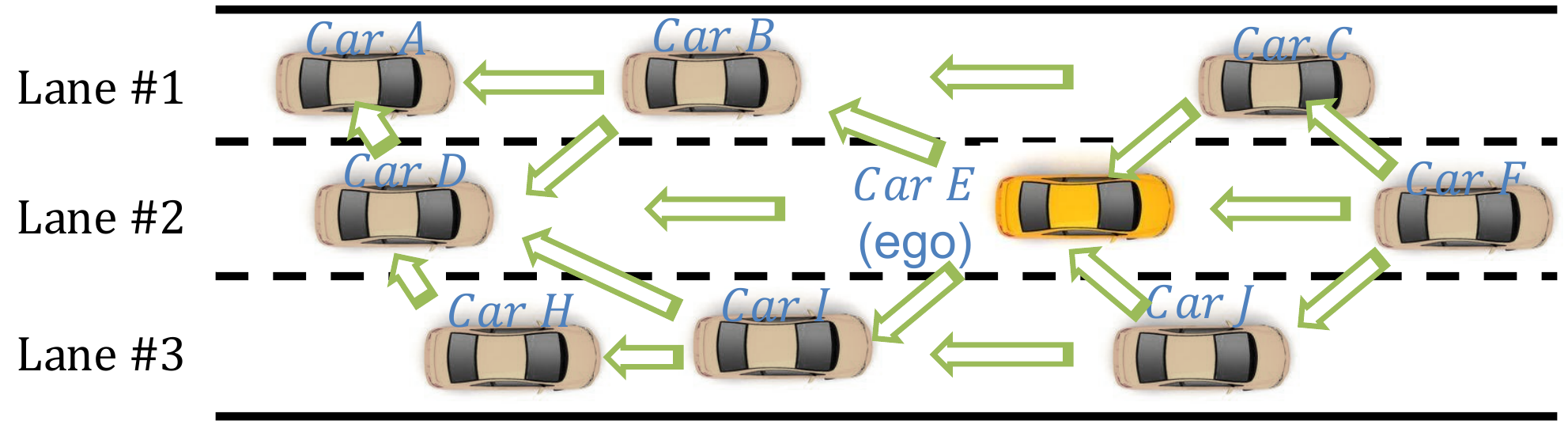}
  \caption{An illustrative scenario of a 3-lane freeway. It shows the vision sharing flow of all the vehicles. There could be human-driven vehicles on street. To make communication relations clear, we only show CAVs here.}\label{fig:image_flow}
\end{figure}



Each vehicle sense its own onboard camera and LIDAR data. 
It is unrealistic to share the raw camera images and point clouds from LIDAR with others due to the following two reasons: (i) the limited bandwidth of a V2V link~\cite{miller2020cooperative}; (ii) the same shared information on surrounding vehicles needs to be repeatedly learned for lane information extraction, resulting in computing resource waste. 
Hence, we first process the vision information locally using CNNs and then share the extracted features to other vehicles. Because the raw images and point clouds need to be synchronously processed and shared for behavior planning, we further develop a weight pruning technique to speed up the slower process. 

\subsubsection{CNN-Driven Shared Vision}
\label{sec:cnn_vision}

\begin{figure}[t] 
\centering \includegraphics[width=\columnwidth]{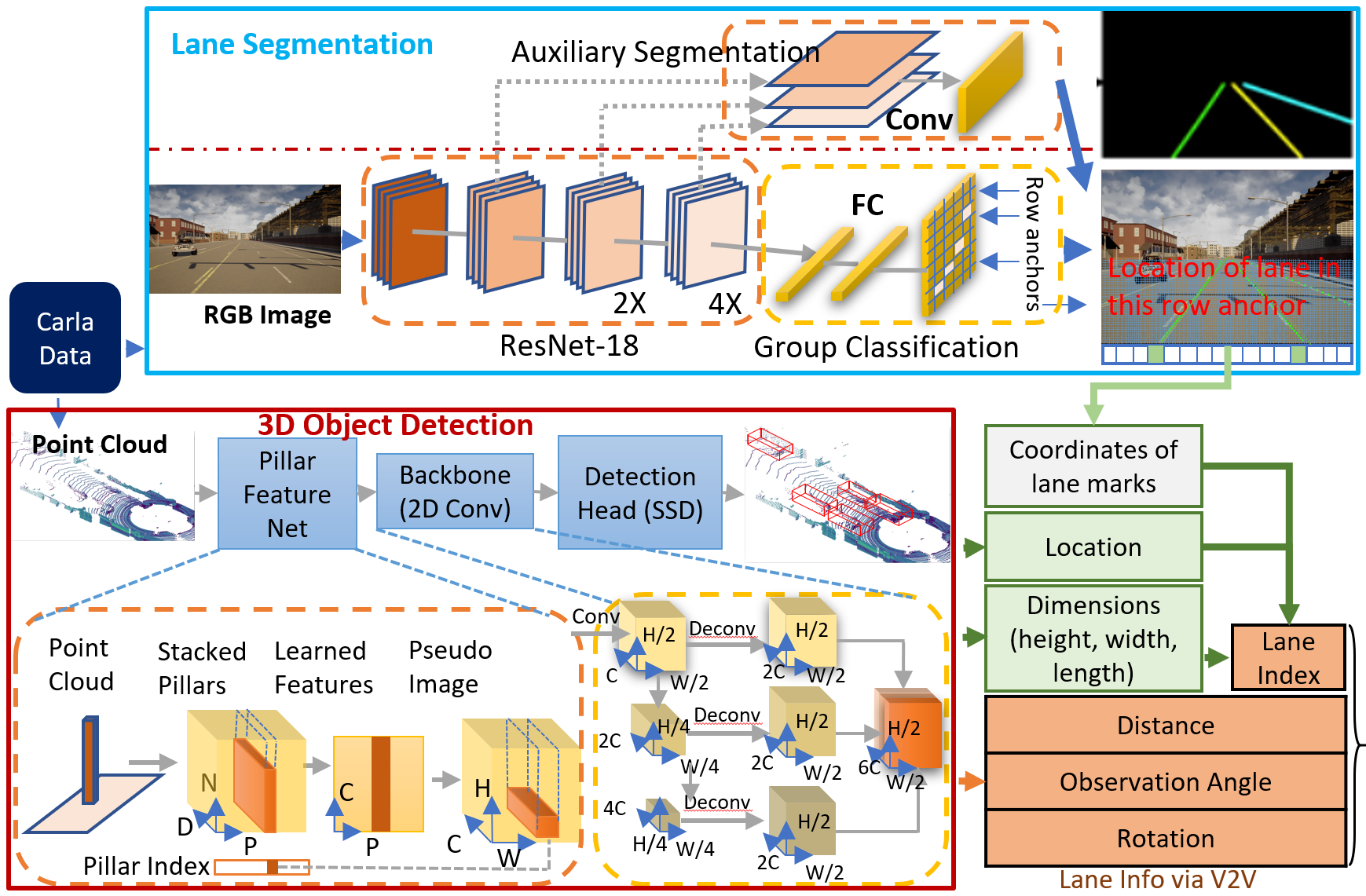}
\vspace{-20pt}
\caption{\label{vision_plot}Detail structure of vision procedure 
}
\vspace{-3pt}
\end{figure}

We develop an end-to-end CNN-driven shared vision solution to derive the relative lane index of surrounding vehicles, i.e., which lane each surrounding vehicle is on. In Fig.~\ref{vision_plot}, 
the shared vision model has two main branches. 
(i) Lane segmentation
for global context and aggregates auxiliary segmentation task to utilize multi-scale features. 
The RGB images from the onboard camera are divided into equal-sized grids, and grids in the same rows are defined as row anchors. Grids with appearance of the lane mark on the row anchor will be colored green. Detailed pixel coordinates of all four-lane marks is generated by the network and saved in a hash table. 
(ii) The 3D OD (object detection) 
uses LIDAR point cloud as input. 
It converts raw point cloud to a sparse pseudo-image by stacking pillars and learning features from it for scatter~\cite{qi2017pointnet}. 
We process the pseudo-image into high-level representation by first continuously down-sampling learned features to small spatial resolution, then up-sampling and concatenating each down-sampled features, to predict 3D bounding boxes for neighbouring vehicles.
The location of vehicles in camera coordinates, dimensions (height, width, length), observation angle, distance, and rotation is generated.

As camera RGB image and LIDAR point cloud are paired under same timestamp, in the second step, we derive the lane index of each surrounding vehicle by fusing the location and dimension of the surrounding  vehicle (obtained by 3D OD) with coordinate of lane marks obtained by lane segmentation. We then send the lane index along with distance, observation angle, and rotation info to the DRL for behavior planning.

\subsubsection{Weight Pruning to Enhance Synchronous Information Share}
We observe that the processing time of point cloud data using PointPillars is significantly longer (11.16$\times$) than that of lane segmentation.
The 3D OD becomes the ``critical path" of the overall vision process. Therefore, we develop a weight pruning technique
on the PointPillars network to reduce the running time. 
Many investigations have shown that there exists redundancy in CNN model parameters~\cite{han2015learning,iandola2016squeezenet,luo2017thinet}. 
CNN {weight pruning} can be used to exploit the redundancy in the parameterization of deep architectures, while maintaining the CNN model accuracy.  We formulate the weight pruning problem in an $N$-layer CNN as:
$\underset{ \{{\bf{W}}_{i}\}}{\text{min}}
~\mathcal{F} \big( \{{\bf{W}}_{i}\}_{i=1}^N\big), \text{subject to}~
{\bf{W}}_{i}\in {\bf{S}}_{i}, \; i = 1, \ldots, N$,
where ${\bf{W}}_{i}$
represents the weights in the $i$-th layer. 
$\mathcal{F}$ is the CNN loss function with respect to ${\bf{W}}_{i}$.
${\bf{S}}_{i}= \{{\bf{W}}_{i}\mid \mathrm{cardinality}({\bf{W}}_{i})\le l_{i}\}$, where $l_{i}$ is desired numbers of non-zero weights. 
Through weight pruning, we bring acceleration in computation and reduction in memory, therefore satisfying the real-time requirement.



\section{Behavior Planning}
\label{sec:learning}


Since autonomous driving is a safety-critical application, the exploration policy in reinforcement learning (RL) for CAVs must be designed rigorously to avoid potential accidents. The action $a$ generated by the traditional $\epsilon$-greedy method~\cite{mnih2015human} may not be safe. 
To ensure there are no collisions during the training process, we add a feedback process to find safe feasible actions. 

\begin{figure}[h]
  \centering
  \includegraphics[width=0.42\textwidth]{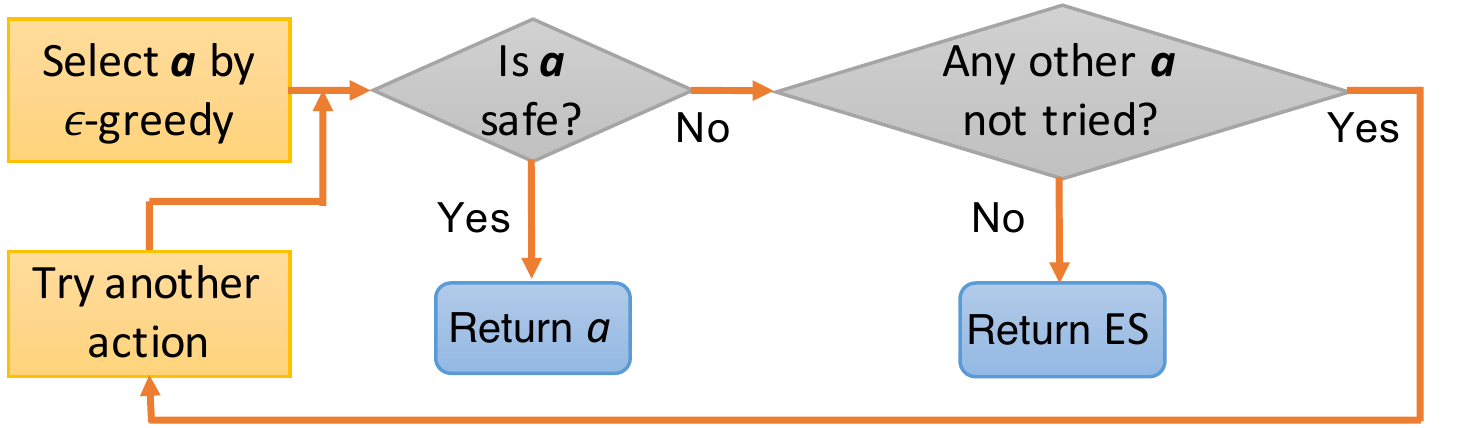}
  \vspace{-10pt}
  \caption{The safe action mapping $\Phi$ to get the safe action. If action $a$ is safe, then return the safe action $a$; if not, then try other actions; if no action is safe, then emergency stop (ES).}\label{fig:feedback}
\end{figure}

\subsubsection{Safe Action Mapping}
\label{sec:feedback_action}
The safe action mapping $\Phi$ to get the safe action is shown in Fig.~\ref{fig:feedback}. Once the RL agent
selects an action, we evaluates whether it is feasible to implement this action by checking safety constraints. If $a$ is safe, then the feedback action $a' = a$; if not, we will search other actions in $\mathcal{A}$ that haven't been tried and find a safe one. If all the actions in $\mathcal{A}$ are not safe in the worst case, then the controller will apply the emergency stop (ES) process. The ES is a special action that is not included in the action set $\mathcal{A}$. It will only be performed in an emergency scenario where all normal actions are not safe. 



\subsubsection{Safe Actor-Critic Algorithm}
\label{sec:Q_learning}

We design a safe actor-critic algorithm as shown in Alg.~\ref{alg:actor-critic} for each CAV to learn a centralized critic $Q$ by minimizing the Bellman loss: 
$
    \mathcal{L}(\theta^i) = \mathbb{E}_{o, a, r, o'}[(y - Q^i(o, a; \theta^i))^2],
$
where $y = r^i + \gamma \cdot Q'^i(o', a'; \theta'^i)|_{a'^{i} = \pi'^{i}(o'^{i})}$ and $Q'^i$ is the target network for the critic, $\pi'^i$ is the target network for the actor. Then we use this critic to train a localized actor $\pi^i(o^i; \theta^i)$ using the gradient
$
\nabla_{\theta^i}J =  \mathbb{E}_{o, a \sim \mathcal{D}} \left[ \nabla_{\theta^i} \pi^i(o^i)\nabla_{a^i}Q^i(o, a) \right].
$
In this algorithm, we use a replay buffer to store the transition experience $(o, a, r, o')$, where $o$ is the current observation, $a$ is the action, $r$ is the reward, and $o'$ is the next observation. The safe action mapping in Fig.~\ref{fig:feedback} assures the policy used to produce a new transition experience is safe. When training the critic, a minibatch is sampled from the replay buffer to decorrelate data.

\begin{figure}[htb]
  \centering
  \includegraphics[width=3.3in]{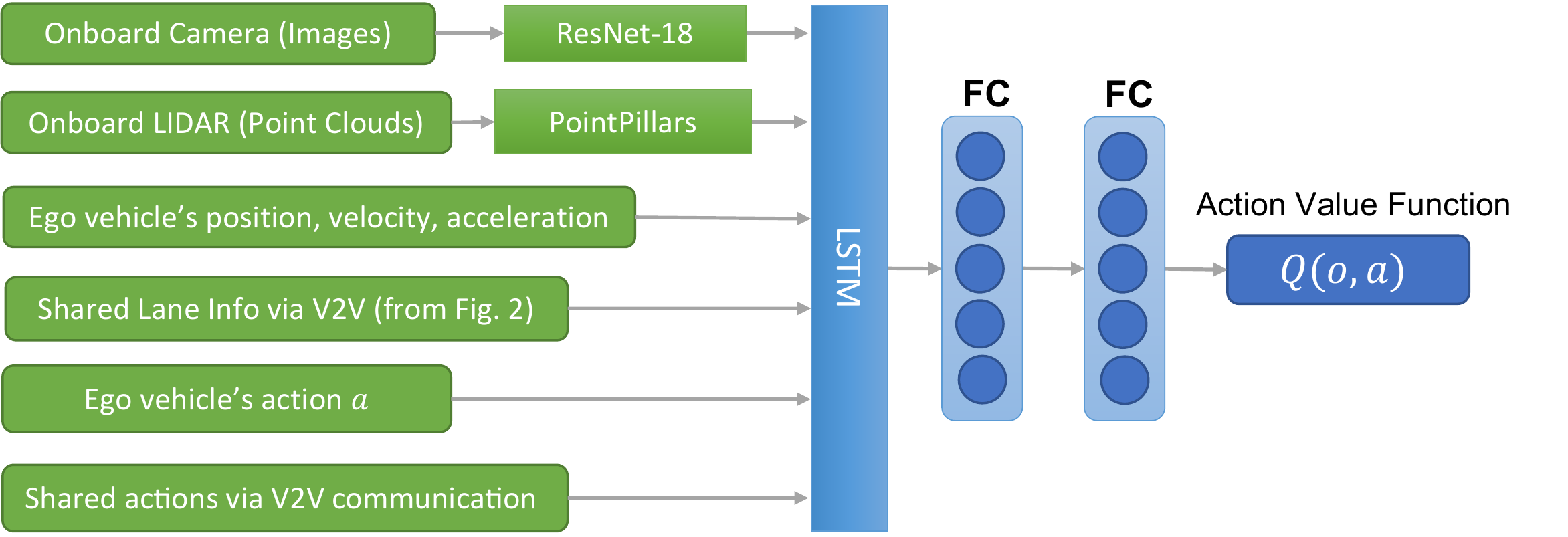}
  \caption{The $Q$-network for the behavior planning with the LSTM layer for the action-observation history. ``FC" represents fully connected layer.}\label{fig:Qnet}
\end{figure}

\begin{algorithm}[h]
\SetAlgoLined
 Randomly initialize the critic network $Q^i$ and the actor network $\pi^i$ for agent $i$. Initialize target networks $Q'^{i}, \pi'^{i}$. Initialize the replay buffer $\mathcal{D}$\;
 \For {each episode}
 {
    Initialize the observation $o$\;
    \For {each timestep}
    {
        With probability $\epsilon$, select action $a^i= \Phi (\pi^i(o^i))$ for each agent $i$, where $\Phi$ is the safe action mapping in Fig.~\ref{fig:feedback}. With probability $1 - \epsilon$, select an action $a^i$ for each agent $i$ in $\{ \Phi(a^i)|a^i \in \mathcal{A} \}$ randomly\;
        
        
        Execute actions $a = (a^1,...,a^n)$ and observe reward $r$ and new observation $o'$\;

        Store $(o, a, r, o')$ in replay buffer $\mathcal{D}$ and set $o \leftarrow o'$\;
        
        \For {each agent $i$}
        {
            
            Sample a random minibatch with size $K$ from $\mathcal{D}$\;
            
            Set $y^i_k = r^i_k + \gamma Q'^{i}(o'_k, a')|_{a'^{i} = \pi'^{i}(o'^{i})}$\;
            
            Update critic by minimizing the loss $\mathcal{L}(\theta^i) = \frac{1}{K}\sum_k (y_{k}^i - Q^{i}(o_k, a_k))^2$\;
            
            Update actor using the gradient $\nabla_{\theta^i}J = \frac{1}{K}\sum_k \nabla_{\theta^i} \pi^i(o^{i}_k)\nabla_{a^i} Q^i(o_k, a_k)$ where $a^i = \pi^i(o^{i})$\;
            
            Update target networks: $\theta'^{i} \leftarrow \tau\theta^i + (1-\tau)\theta'^{i}$.
        }
    }
 }
 \caption{Safe Actor-Critic Algorithm}
 \label{alg:actor-critic}
\end{algorithm}
\setlength{\textfloatsep}{1pt}

\subsubsection{$Q$-network}
The $Q$-network used in Alg.~\ref{alg:actor-critic} is shown in Fig.~\ref{fig:Qnet}. We consider the action set $\mathcal{A}$ includes \{Keep Lane (KL), Change Left (CL), and Change Right (CR)\}. The observation for each CAV includes: (i) onboard camera, LIDAR; (ii) position, velocity, and acceleration; (iii) processed shared vision information and actions of its immediate leaders on its current and neighbor lanes via V2V.  Our algorithm also works in mixed traffic with both autonomous and human-driven vehicles. When there is no CAV in the surrounding vehicles, the ego vehicle can always use its onboard sensors' information as the input of the $Q$-network. 

\subsubsection{Reward Function}
\label{sec:reward}
We consider two system-level reward as evaluation criteria: average velocity $\bar{v}$ and average comfort $\bar{c}$. The comfort of single vehicle (for passenger's experience) is defined based on its acceleration and action as follows:
\begin{equation}
\label{equ:comfort}
\small
    c(\dot{v}, a) = \begin{cases} 
			 3, & \text{if } \lvert \dot{v} \rvert< \Theta \text{ and } a = KL; \\
			2, & \text{if } \lvert \dot{v} \rvert \geq \Theta \text{ and } a = KL; \\
			 1, & \text{if } a = CL/CR; \\
			 0, & \text{if in } ES. \end{cases} 
\end{equation}
where $\dot{v}$ is the acceleration, $a \in \mathcal{A}$ is the behavior action, $\Theta$ is a predefined threshold. The reward function is defined as:
\begin{equation}
\mathbf{r}(s, a) = w \cdot \bar{v} + \bar{c},
\label{equ:reward_centralized}
\end{equation}
where $s \in \mathcal{S}$, and $w$ is a trade-off weight.

\section{Experiment}
\label{sec:experiment} 


We use CARLA~\cite{Dosovitskiy17}, an open source simulator which supports development, training, and validation of autonomous driving systems, to validate our proposed method. Each CAV is integrated with a camera sensor for capturing RGB images, and a LIDAR sensor for generating point clouds. The resolution of camera image is $375\times1,242$ pixels. We set row anchors ranging from 100 to 370 with intervals of 10, and the number of grids is 100. We scale each image to $288\times800$. Each point cloud from LIDAR is stored with the 3 coordinates (the ego vehicle being the origin), representing forward, left, and up respectively, and an additional reflectance value. To augment the dataset, we apply random mirroring flip along the forward axis, and a global rotation and scaling. In evaluation, we set forward and up axes' resolution to $0.16 m$, max number of pillars to 12,000, and max number of points per pillar to 100. Images and point clouds are paired under the same timestamp and are processed jointly under the CNN.

We assume all CAVs to share their processed vision information that includes features \{lane index, distance, observation angle, rotation\} of neighboring vehicles  with their immediate followers as introduced in Section~\ref{sec:info_sharing}. 
Each CAV collects transition experience and store them in the replay buffer. Then they use minibatch gradient descent to learn the $Q$-function. The learning method detail is introduced in Section~\ref{sec:learning}. The hyperparameters of our Alg.~\ref{alg:actor-critic} is shown in Table~\ref{tbl:hyperparameter}.

\begin{table}[h!]
\centering

\begin{tabular}{l|c}
\toprule
\textbf{Parameter}                         & \textbf{Value}\\ 
\midrule
optimizer                         & Adam                  \\ 
learning rate                     & 0.01                  \\ 
discount factor                   & 0.9                  \\ 
replay buffer size                & $10^6$                  \\ 
hidden size in FC and LSTM layers    & 128                    \\ 
minibatch size  & 64                 \\ 
activation function                      & ReLU                  \\ 
frequency to update target network     & 100                  \\ 
\bottomrule
\end{tabular}
\caption{Safe Actor-Critic Hyperparameters.}
\label{tbl:hyperparameter}
\end{table}

The host machine adopted in our experiments is a server configured with Intel Core i9-10900X processors and four NVIDIA RTX2080Ti GPUs. Our experiments are performed on Python 3.7.6, GCC7 7.5, PyTorch 1.6.0, and CUDA 11.0.  

\subsection{CNN-Driven Shared Vision}
\label{sec:vision_experiment}

 For evaluation, PointPillars uses mean average precision (mAP) as standard, while Ultra-Fast-Lane-Detection uses ``accuracy", which is calculated as: $\frac{\sum_{\text {clip}} C_{\text {clip}}}{\sum_{\text {clip}} S_{\text {clip}}}$, where $C_{clip}$ is the number of lane points predicted correctly and $S_{clip}$ is the total number of ground truth in each clip. We show the performance (accuracy or mAP) and running speed in Table~\ref{vision_result}. 
{The lane segmentation achieves high accuracy with low latency. It can process 313 images in one second. The 3D object detection has the running time of $28$ images per second (img/s), which means 100 images can be processed in around 3.6 seconds.}
Our weight pruning technique significantly increases the speed  (by $3.5\times$) with a very small accuracy degradation. {It decreases the processing time of 100 images from 3.6 seconds to one second.} Overall, for the parallel lane segmentation and 3D object detection, our weight pruning technique significantly speeds up the CNN-driven shared vision process, satisfying real-time processing requirements.
Our vision processing method can be used to mixed traffic setting that includes human-driven vehicles. The lane segmentation and object detection is the same when dealing with both autonomous and human-driven vehicles.

\begin{table}[h]
\centering
\begin{tabular}{c|cc}
\toprule
CNN Model                  & performance (\%) & speed (img/s) \\
\midrule
Lane       & \multirow{2}{*}{accuracy: 95.82} & \multirow{2}{*}{313} \\
Segmentation  & & \\
\midrule
3D object     & \multirow{2}{*}{mAP: 76.5}       & \multirow{2}{*}{28} \\
detection (OD) & & \\
\midrule
3D OD after  & \multirow{2}{*}{mAP: 73.6}       & \multirow{2}{*}{99} \\
weight pruning & & \\
\bottomrule
\end{tabular}
\caption{Summary of Result From Vision Network.}
\label{vision_result}
\end{table}

\subsection{System Efficiency Improvement}

In this section, we show our algorithm improves the CAVs' system efficiency in terms of the average velocity and the average comfort as defined in the reward function~\eqref{equ:reward_centralized}. 

\subsubsection{Comparison Under Different CAV Ratios}

Our approach improves the average velocity and the average comfort as the CAV ratios (the total CAV number divided by the total number of all vehicles) get higher. In this set of experiments, the total number of CAVs ranges from 0 to 30 as listed in Table~\ref{tbl:mixed_setting}. We compare the average velocity and comfort for all vehicles under different CAV ratios. The comfort of a single vehicle is defined in Eq.~\eqref{equ:comfort}. It is averaged among all the vehicles as one criterion. The velocity and comfort are averaged over all the 40000 timesteps used in the simulation. The result of our approach is shown in Table~\ref{tbl:mixed_setting}. All CAVs use our safe actor-critic Alg.~\ref{alg:actor-critic} introduced in Section~\ref{sec:learning}. The result in Table~\ref{tbl:mixed_setting} shows the average velocity and comfort of the entire mixed traffic. We use CARLA's built-in human-driven vehicle in the mixed traffic~\cite{Dosovitskiy17}. In the result of Table~\ref{tbl:mixed_setting}, the average velocity and comfort increase when the CAV ratio gets higher. This gives us insights that the penetration of the CAVs can improve traffic efficiency in the future. 

\begin{table}[h]
  \centering
  \begin{tabular}{c|cc|cc}
  \toprule
  CAV & CAV  & HDV & average  & average  \\
  ratio & number & number & velocity (mph) & comfort\\
  \midrule
  0 & 0 & 30 & 60.06 & 2.61 \\
  0.17 & 5 & 25 & 61.82 & 2.64 \\
  0.33 & 10 & 20 & 64.70 & 2.68 \\
  0.5 & 15 & 15 & 65.14 & 2.72 \\
  0.67 & 20 & 10 & 65.18 & 2.74 \\
  0.83 & 25 & 5 & 65.53 & 2.77 \\
  1 & 30 & 0 & 66.15 & 2.81 \\
  \bottomrule
  \end{tabular}
  
  \caption{The system efficiency comparison under different CAV ratios. Our approach improves the average velocity and the average comfort as the CAV ratios get higher. (HDV: Human-driven vehicles)}
  \label{tbl:mixed_setting}
\end{table}

\subsubsection{Comparison Under Different Traffic Densities}

\begin{figure}[htb]
  \centering
  \includegraphics[width=\columnwidth]{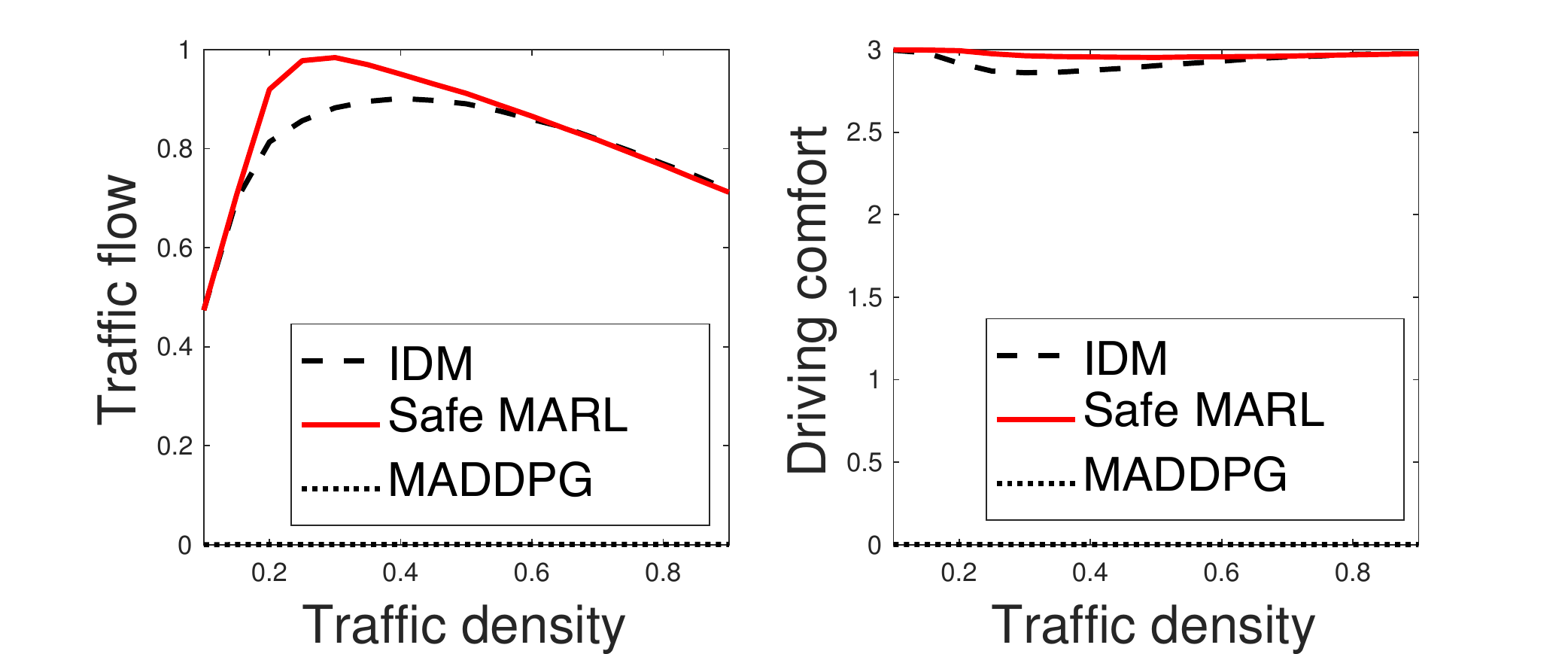}
  \caption{Our approach (safe MARL) improves the traffic flow and driving comfort under different traffic densities.}\label{fig:idm}
\end{figure}

Our approach improves the traffic flow and average comfort under different traffic densities. The traffic density $\rho$ is the ratio between the total number of vehicles and the road length. We compare the traffic flow and average comfort under different traffic densities among our safe MARL approach using Alg.~\ref{alg:actor-critic}, the MADDPG~\cite{lowe2017multi} algorithm and an intelligent driving model (IDM)~\cite{talebpour2016influence}. The traffic flow reflects the quality of the road throughout with respect to the traffic density. It is calculated as $\rho \times \bar{v},$ where $\bar{v}$ is the average velocity of all the vehicles \cite{rios2018impact}. The IDM is a common baseline in autonomous driving. In IDM, the vehicle's acceleration is a function of its current speed, current and desired spacing, and the leading and following vehicles' speed~\cite{talebpour2016influence}. Building on top of these IDM agents, we add lane-changing functionality using the gap acceptance method in~\cite{butakov2014personalized}. In this set of experiments, we keep CAV ratios at 0.6. As shown in Fig.~\ref{fig:idm}, the safe MARL agent gets both larger traffic flow and better driving comfort when traffic density $\rho$ is low. When $\rho$ grows, the result of the safe MARL agent gets worse, but it is still comparable with the IDM. When the road is saturated, lane-changing tends to downgrade passengers' comfort but cannot bring higher speed. Consequently, a better choice is to keep lanes when $\rho$ is high, and there is no significant difference between the safe MARL and the IDM. We also add the result using MADDPG in Fig.~\ref{fig:idm}. Both the traffic flow and the driving comfort are very small (a positive number but close to 0) using MADDPG, and many collisions occurred during the simulation. This is because the MADDPG does not have a safety guarantee.

\subsection{Safety Guarantee}

In this section, we show our algorithm has a safety guarantee. We compare our safe actor-critic algorithm with the MADDPG~\cite{lowe2017multi} algorithm. We assign a negative reward in MADDPG for each collision. Our approach avoids the execution of unsafe actions that can lead to collisions. Table~\ref{tbl:number_feedback} shows the total number of unsafe actions executed by the safe MARL and MADDPG under different traffic densities. The traffic density $\rho$ is the ratio between the total number of vehicles and the road length. The number in Table~\ref{tbl:number_feedback} is averaged over the last 10 episodes, which has a maximum timestep of 40000. When $\rho=0.9$, our approach has 0 unsafe actions while the MADDPG has 242976 unsafe actions because it does not have a safety module.

\begin{table}[h]
  \centering
  \begin{tabular}{c|cccc}
  \toprule
  $\rho$ & 0.1 & 0.2 & 0.3 & 0.4  \\
  \hline
  Safe MARL & 0 & 0 & 0 & 0  \\
  \hline
  MADDPG & 4416 & 9510 & 21897 & 43491 \\
  \midrule
  $\rho$ & 0.5 &  0.6 & 0.7 & 0.8  \\
  \hline
  Safe MARL & 0&  0 & 0 & 0  \\
  \hline
  MADDPG &61689 & 91154 & 133135 & 191404 \\
  \bottomrule
  \end{tabular}
  
  \caption{Total number of unsafe actions executed by the safe MARL and MADDPG under different traffic density $\rho$. Our approach has 0 unsafe action by using the safe action mapping.}
  \label{tbl:number_feedback}
\end{table}

\begin{figure}[h]
  \centering
  \includegraphics[width=3.5in]{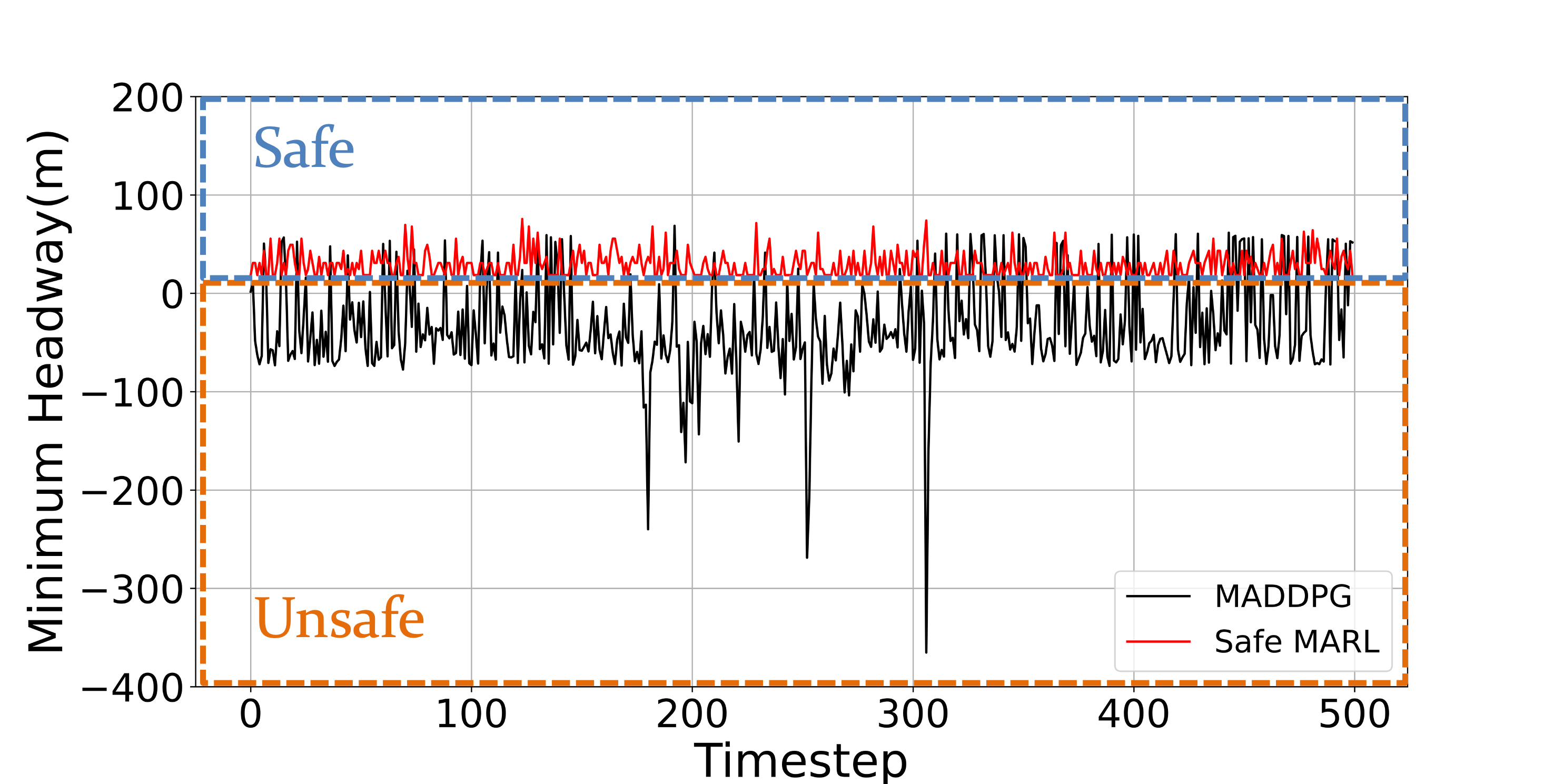}
  \vspace{-10pt}
  \caption{Our approach (Safe MARL) can maintain a safe headway with neighboring vehicles while the MADDPG cannot. The figure shows the minimum headway across all the vehicles in the first 500 timesteps of one episode when $\rho = 0.6$.}\label{fig:headway}
\end{figure}

Our approach can maintain a safe headway with neighboring vehicles while the MADDPG cannot. The headway is the distance between two consecutive vehicles following each other. In Fig.~\ref{fig:headway}, the minimum headway across all the vehicles is shown in the first 500 timesteps of one episode when $\rho = 0.6$. We see that the headway is always greater than 0 using our safe actor-critic algorithm with a minimum value of 18.5 meters. Nevertheless, the MADDPG is likely to have a negative headway, which means collisions in reality. Note that we set the minimum car-following distance of CAVs to be 18.5 meters following the study of the safe car-following distance of autonomous vehicles~\cite{arechiga2019specifying}, but this value can be set differently to satisfy the requirements in different scenarios.

\begin{figure}
    \centering
    \subfigure{
    \begin{minipage}{1.5in}
    \centering
    \includegraphics[scale=0.14]{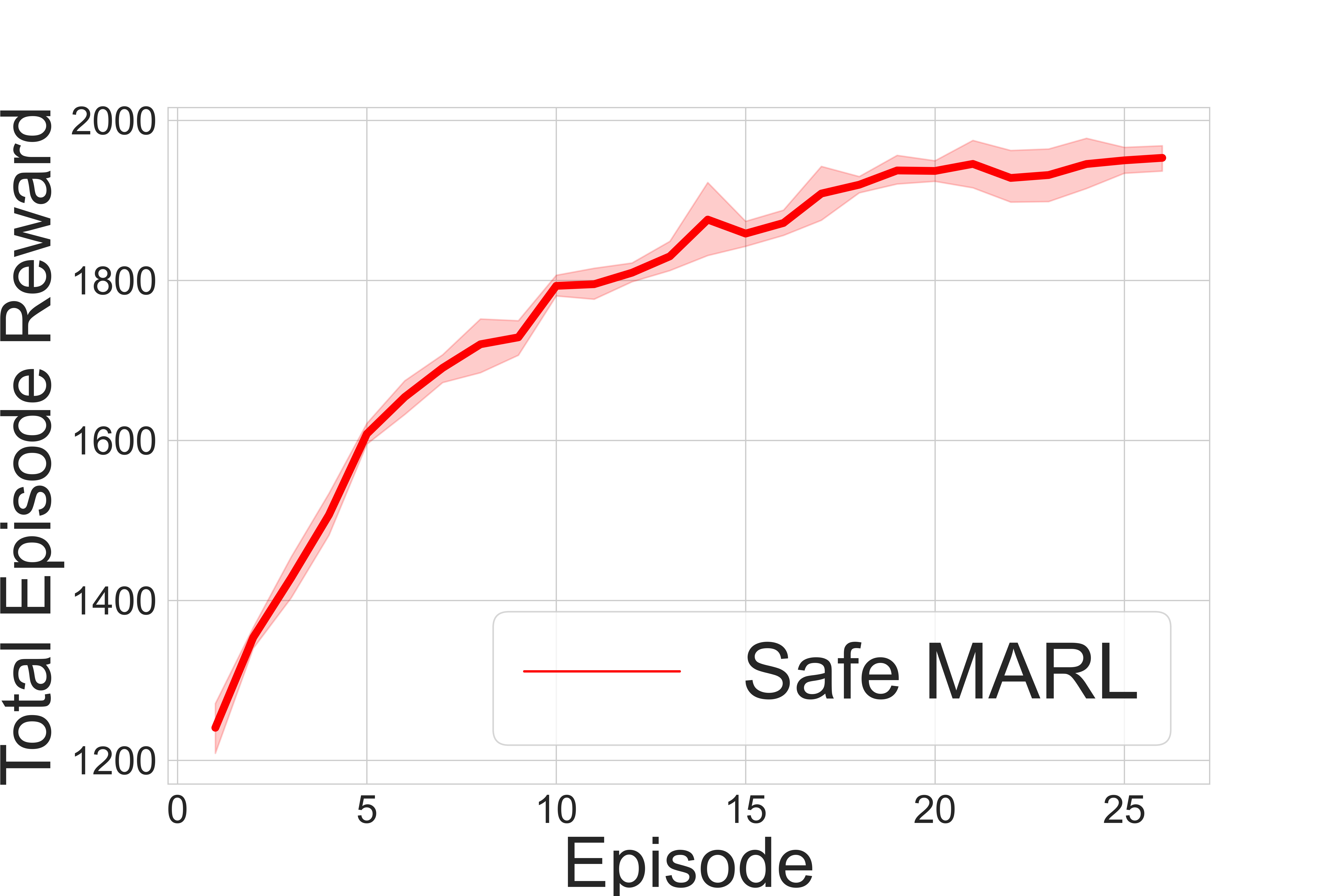}
    \end{minipage}
    }
    \subfigure{
    \begin{minipage}{1.5in}
    \centering
    \includegraphics[scale=0.14]{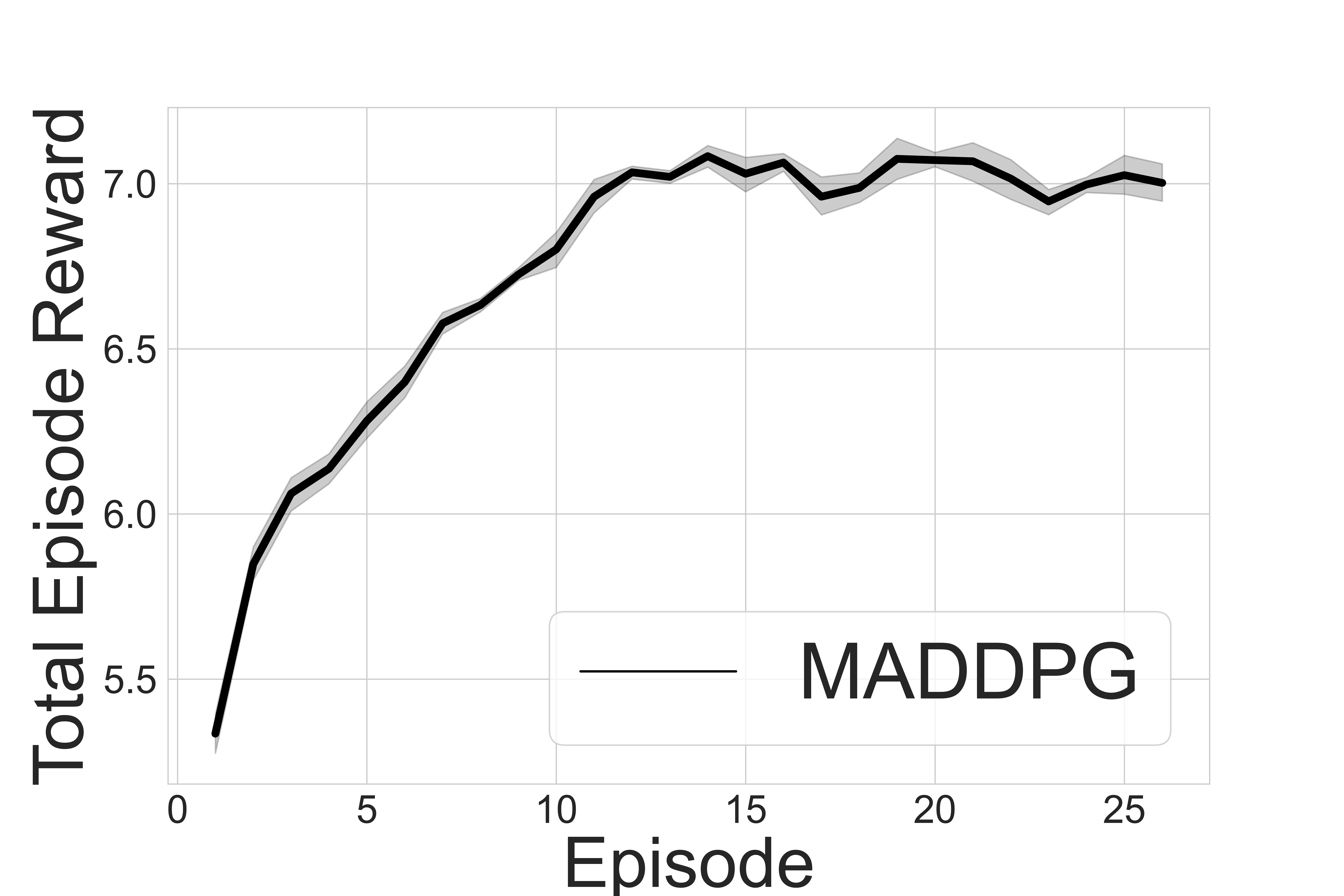}
    \end{minipage}
    }
    \caption{Our approach (safe MARL) gets a higher total episode reward compared to the MADDPG during the training process, because our approach can guarantee a safe training process and run for the maximum episode length.}\label{fig:reward_result}
\end{figure}

Our approach gets a much larger total episode reward than the MADDPG. The total episode reward is the summation of all stage-wise rewards defined in Eq.~\ref{equ:reward_centralized} for each episode. As shown on the left of Fig.~\ref{fig:reward_result}, the maximum total episode reward using our approach is about 1940, which is firstly reached in the 20th episode. In the right figure, the maximum total episode reward using the MADDPG is about 7, because some collision terminates the episode. They have different initial rewards because the neural networks are randomly initialized and the action is selected by the $\epsilon$-greedy method with randomness. Our approach runs about 30 minutes in each episode as it has a safety guarantee and runs for the maximum episode length. Yet, each episode stops quickly in about 5 seconds using the MADDPG.

\subsection{Obstacle-At-Corner Scenario and Benefit of Shared Vision}

\begin{figure}[h]
  \centering
  \includegraphics[width=3in]{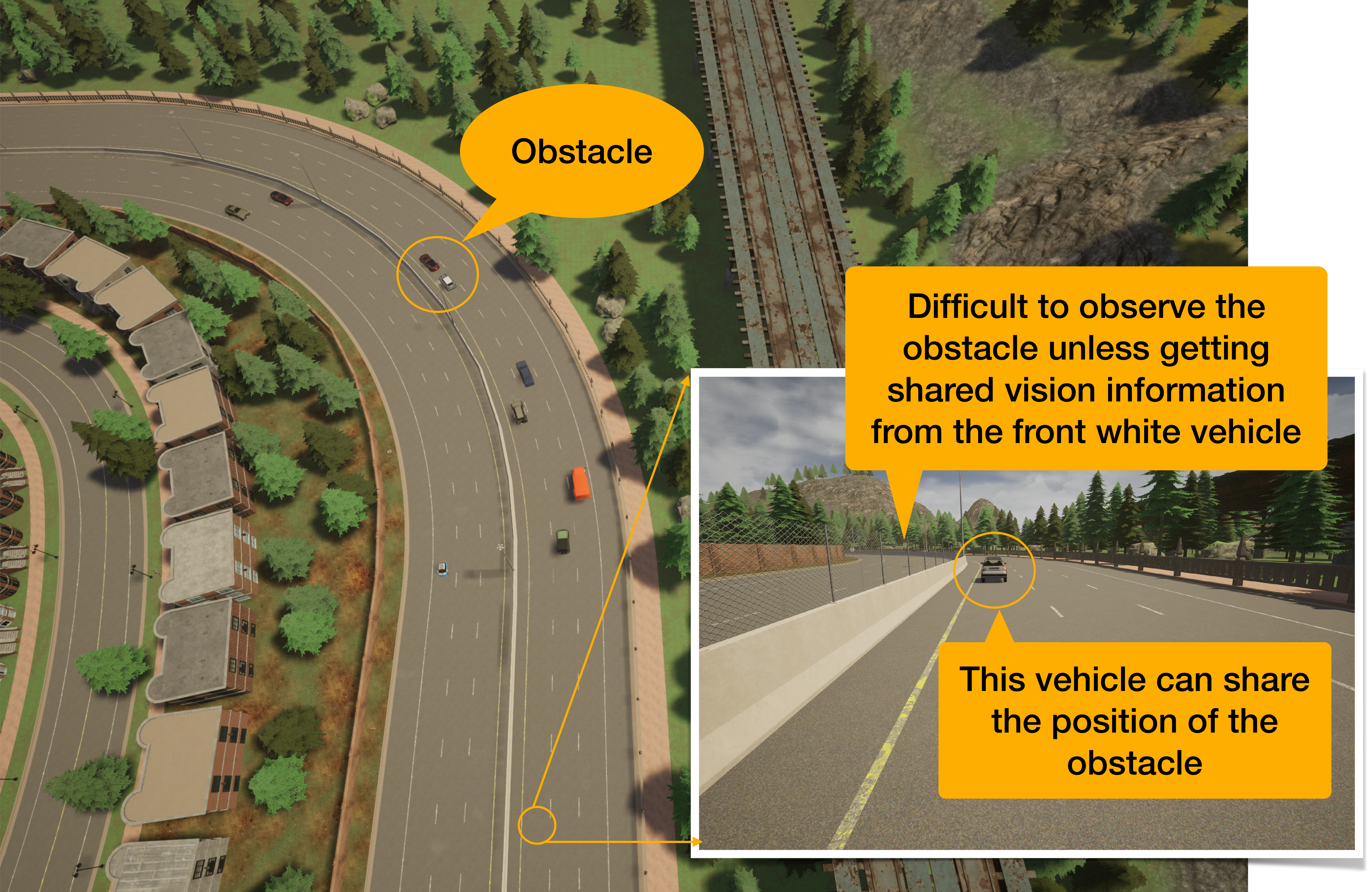}
  \caption{The obstacle-at-corner scenario where there are obstacles in a left-turning corner. The vehicles on the road are autonomous vehicles. The coming vehicles' view is blocked such that they cannot observe the obstacles.}\label{fig:obstacle}
\end{figure}

We construct a scenario called obstacle-at-corner to show how sharing vision information can help autonomous vehicles make wise lane-changing decisions ahead of time. As shown in Fig.~\ref{fig:obstacle}, there are obstacles at a left-turning corner (represented by two stationary vehicles). The right bottom figure shows the view of a vehicle that comes in the direction of this curve road. It is quite difficult to observe the obstacles merely relying on its own sensors. In this case, if the white front vehicle can share its observation, the coming vehicle can get to know there are obstacles before entering the turning corner.

\begin{figure}[h]
  \centering
  \includegraphics[width=3.2in]{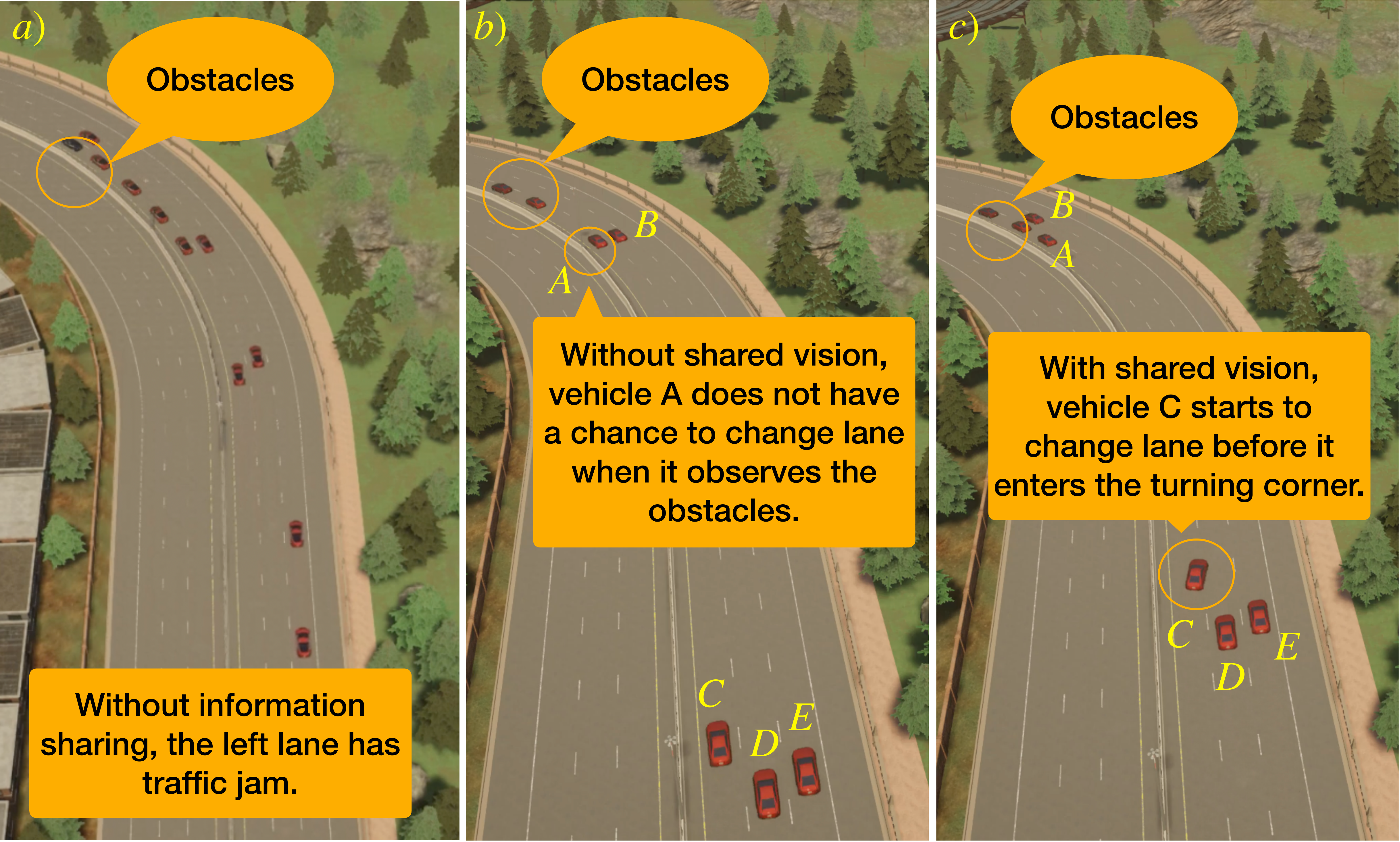}
  \caption{Without any information sharing, there is a traffic jam in the left lane. With shared vision from $A$ or $B$, using our safe MARL policy, vehicle $C$ can change its lane before it enters the left-turning corner.}\label{fig:lanechanging_corner}
\end{figure}

Shared vision with our proposed safe MARL algorithm can help CAVs avoid traffic jams. We test our learned policy in the obstacle-at-corner scenario. When no vehicle can share information in subfigure (a) of Fig.~\ref{fig:lanechanging_corner}, there are more vehicles blocked on the left-most lane and causing a traffic jam. Subfigures (b) and (c) are the screenshots taken in two close timesteps. In subfigure (b), vehicle $A$ is blocked in the left lane because there is vehicle $B$ in the middle lane and $A$ does not realize there are obstacles before it enters the corner (there is no neighboring vehicle that can share the obstacle information in advance in this scene;  two obstacles are not CAVs). In subfigure (c), the coming vehicle $C$ in the left-most lane starts to change to the middle lane before it observes the obstacles because it gets the shared vision information either from $A$ or $B$. We use the safe action mapping introduced in Sec.~\ref{sec:learning} to ensure the safety of this lane-changing.

\section{Conclusion}
\label{sec:conclusion}
This paper studies how to solve behavior planning challenges for CAVs, that is, how to utilize V2V communication to improve traffic efficiency while also following safety requirements. We design an integrated information sharing and safe multi-agent reinforcement learning framework, which utilizes local observations and shared information in order to safely explore a behavior planning policy. The safe action mapping guarantees the safety of the training and execution process of the proposed MARL framework.
We conduct the CAV simulation in CARLA. In the experiment, our weight pruning technique increases the speed of the 3D object detection (OD) by 3.5$\times$ with small accuracy degradation,
since OD is the bottleneck for sharing vision. We also show that our approach improves the average velocity and average comfort under different CAV ratios and different traffic densities. Our approach avoids unsafe actions and maintains a safe distance from neighboring vehicles. We also construct the obstacle-at-corner scenario to show that the shared vision can help vehicles to avoid traffic jams. It is considered as future work to improve the robustness of the MARL policy under state uncertainties that are caused by V2V communication or sensor measurement errors.

\bibliography{aaai23}

\end{document}

%% file: defs.tex
\newif\ifuseboldmathops
\newif\ifuseittextabbrevs
\useboldmathopstrue   

\ifuseittextabbrevs

\else

\fi

\ifuseboldmathops

\else

\fi

\ifuseboldmathops

\else

\fi

\ifuseboldmathops

\else

\fi

\ifuseboldmathops

\else

\fi

\newcommand{\norm}[1]{\lVert#1\rVert}

\newcommand{\calF}{\mathcal{F}}

\acrodef{mdp}[MDP]{Markov decision process}
\acrodef{dfa}[DFA]{deterministic finite-state automaton}
\acrodef{ltl}[LTL]{linear temporal logic}
\acrodef{ltlf}[LTL$(\calF)$]{quantitative linear temporal logic}
\acrodef{ag}[AG]{Assume-Guarantee}
\acrodef{ssp}[SSP]{Stochastic Shortest Path}
\acrodef{mcmc}[mcmc]{Monte Carlo Markov chain}

\theoremstyle{definition}
 \newtheorem{definition}{Definition}
 \newtheorem{example}{Example}

\acrodef{ltl}[LTL]{linear temporal logic formula}
\acrodef{mdp}[MDP]{Markov decision process}
\acrodef{smdp}[Semi-MDP]{Semi-Markov decision process}

\acrodef{scltl}[scLTL]{syntactically co-safe LTL}